# OpenRad: a Curated Repository of Open-access AI models for Radiology

Original Research

Konstantinos Vrettos[1], Galini Papadaki[1], Emmanouil Brilakis[1], Matthaios Triantafyllou[1], Dimitrios Leventis[1], Despina Staraki[1], Maria Mavroforou[1], Eleftherios Tzanis[1,2], Konstantina Giouroukou[1], Michail E. Klontzas[1,2,#]

1. Artificial Intelligence and Translational Imaging (ATI) Lab, Department of Radiology, School of Medicine, University of Crete, Heraklion, Crete, Greece
2. Division of Radiology, Department of Clinical Science Intervention and Technology (CLINTEC), Karolinska Institute, Huddinge, Sweden

#Corresponding author

***Address for correspondence***
**Michail E. Klontzas**, MD, PhD
Assistant Professor of Radiology
Artificial Intelligence and Translational Imaging (ATI) Lab
Department of Radiology, School of Medicine,
University of Crete
Voutes, 71003, Heraklion, Crete, Greece
Tel: +30 2811391351 E-mail: miklontzas@gmail.com; miklontzas@ics.forth.gr; miklontzas@uoc.gr
ORCID: 0000-0003-2731-933X

## Abstract

The rapid developments in artificial intelligence (AI) research in radiology have produced numerous models that are scattered across various platforms and sources, limiting discoverability, reproducibility and clinical translation. Herein, OpenRad was created, a curated, standardized, open-access repository that aggregates radiology AI models and providing details such as the availability of pretrained weights and interactive applications. Retrospective analysis of peer reviewed literature and preprints indexed in PubMed, arXiv and Scopus was performed until Dec 2025 (n = 5239 records). Model records were generated using a locally hosted LLM (gpt-oss:120b), based on the RSNA AI Roadmap JSON schema, and manually verified by ten expert reviewers. Stability of LLM outputs was assessed on 225 randomly selected papers using text similarity metrics. A total of 1694 articles were included after review. Included models span all imaging modalities (CT, MRI, X-ray, US) and radiology subspecialties. Automated extraction demonstrated high stability for structured fields (Levenshtein ratio > 90%), with 78.5% of record edits being characterized as minor during expert review. Statistical analysis of the repository revealed CNN and transformer architectures as dominant, while MRI was the most commonly used modality (in 621 neuroradiology AI models). Research output was mostly concentrated in China and the United States. The OpenRad web interface enables model discovery via keyword search and filters for modality, subspecialty, intended use, verification status and demo availability, alongside live statistics. The community can contribute new models through a dedicated portal. OpenRad contains approx. 1700 open access, curated radiology AI models with standardized metadata, supplemented with analysis of code repositories, thereby creating a comprehensive, searchable resource for the radiology community.

## 1.Introduction

In the recent years the radiology community has witnessed an unprecedented surge of artificial-intelligence (AI) research (1). Numerous deep-learning models for report writing (2), lesion detection (3), segmentation (4) and many other tasks have been reported in conferences, journals and pre-print servers. Yet, the very abundance of these methods has created a new bottleneck: discoverability, reproducibility and clinical translation are hampered by a fragmented landscape of model distribution.

Published models are scattered across a variety of sources including supplementary material, personal GitHub repositories, institutional webpages or proprietary platforms, forcing users to conduct exhaustive manual searches to locate a model that matches their specific requirement. In addition, the lack of standardized metadata makes it challenging to assess whether a model is appropriate for a given clinical setting or to compare it fairly against alternative solutions (5). The first step to address this issue was the introduction of reporting guidelines for papers relevant to AI models in radiology (6,7). Expanding on this approach, RSNA proposed the AI Roadmap, which suggests standardized ontology terms (RadLex) and format, for reporting AI model data in radiology that disclose details such as intended use, model structure and licensing information (8,9). Nonetheless, adherence to these standards is not widespread, which limits the availability of structured, machine-readable model records. Existing AI model repositories (10–14) are based on model details provided by authors, increasing research transparency. Nonetheless, these works only include a limited handpicked number of model cards, some of which refer to datasets without accompanying models. Importantly, they include manuscripts without open access code that are not reproducible and may not be useful to researchers attempting to develop new models or to apply existing models to enhance radiology workflows.

The aim of this work was to create OpenRad, a comprehensive, up-to-date, searchable public repository of open access radiology AI models, that provide open-access code or model for public use. The proposed work offers: (i) an expert-curated collection of 1694 models spanning all imaging modalities and radiology subspecialties, (ii) standardized model records using the ROADMAP ontology (RadLex terms), ensuring consistent and interoperable reporting, (iii) additional information derived from the analysis of code repositories, such as the availability of trained models or interactive implementations, (iv) option for users to submit new models following the standardized format, (v) live statistical data, based on the included models, capturing the current state of open-access AI research in radiology. This collection is accessible through a Github-based web portal that enables model discovery for researchers, clinicians and industry partners. This repository will render existing algorithms discoverable, facilitating their research use and the translation of promising algorithms into clinically useful tools.

## 2. Materials and Methods

### 2.1 Initial dataset creation

A comprehensive corpus of radiology AI papers was formed by scraping PubMed, arXiv and Scopus using the queries in Supplementary Table 1 and including works published until 12/2025. This search retrieved 5239 records that span medical literature but also computer science journals and conference proceedings which carry a large volume of AI models. Duplicates across databases were removed using the DOI (985 duplicates) and manual screening excluded papers that were not focused on a radiology AI model (1832 papers). Importantly, papers that did not provide open access code/model (704) or referred to datasets (24) were subsequently excluded, yielding a final dataset of 1694 unique articles (Figure 1), leading to a repository where all presented models are accompanied by ready-to-use code/model.

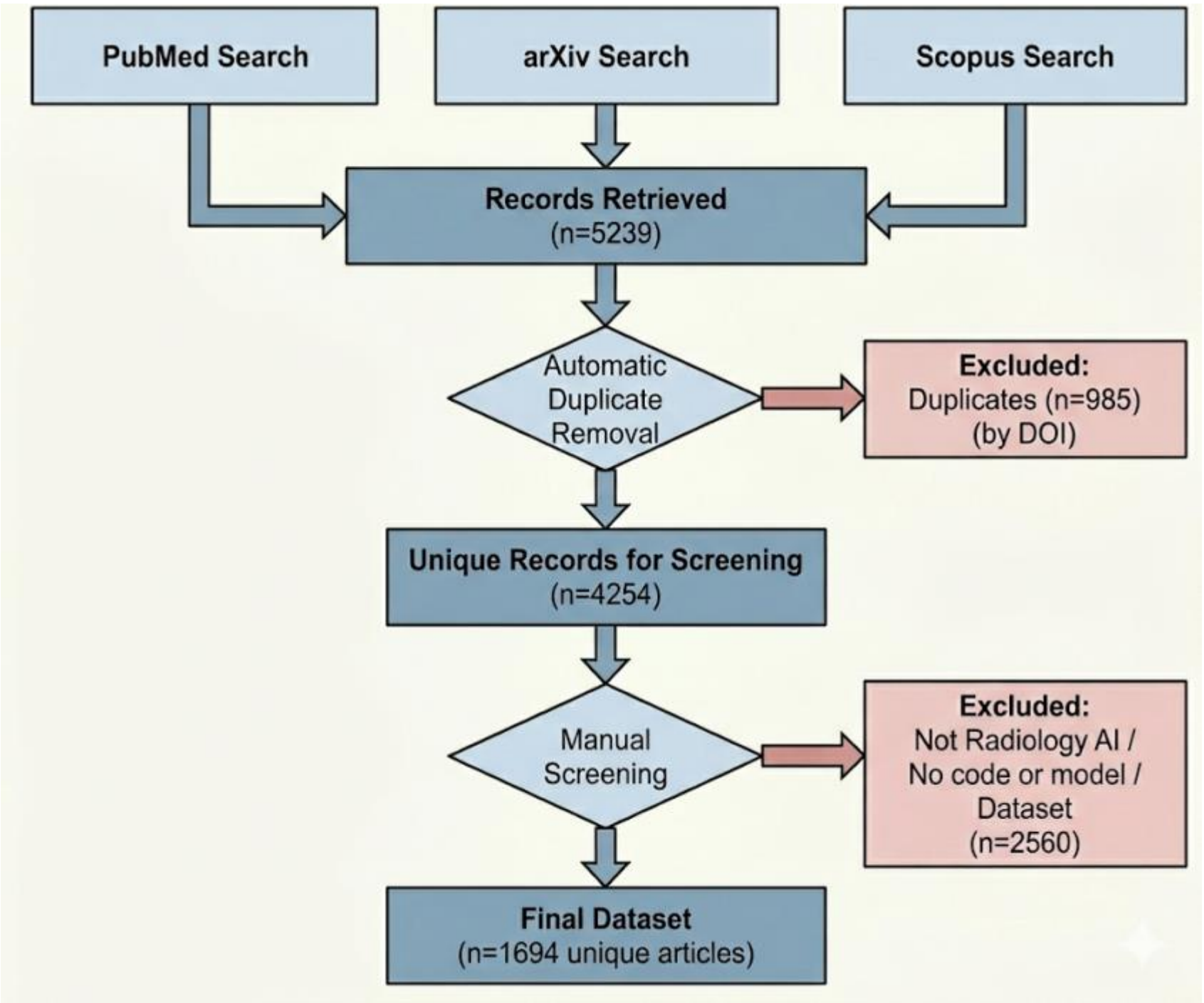

**Fig.1** *Flowchart of the article selection process*

To transform this corpus into a standardized collection of models, the RSNA AI Roadmap reference schema (https://github.com/RSNA/ATLAS/blob/main/model.json) was followed. To enable the creation of model records for thousands of papers, an automated extraction pipeline was devised, which leverages the open-source large language model, gpt-oss:120b, run locally via Ollama. Python 3.10 with the "instructor" library was used to issue structured prompts that populate the selected fields of the RSNA JSON schema. When a full-text PDF was available, it was supplied to the LLM, otherwise the abstract was used. In addition to the recommended RSNA fields (modality, clinical subspecialty, intended use, architecture, performance metrics), the template was enriched with attributes regarding dataset details and validation strategy. The generated model records also include an assessment of the contents of the GitHub repository, if one is provided. This was carried out by automatically querying all URLs identified as code repositories and identifying the presence of pretrained weights or ready-to-use implementations (demos) of the models. The final JSON files contained bibliographic metadata and the extracted technical details.

### 2.2 LLM-generated model record assessment: stability and ground-truth comparison

To evaluate the reliability of the automatically generated records, the intra-LLM stability was checked. A subset of 225 articles was repeatedly processed, with the first generation serving as the reference. The stability was evaluated across key fields such as title, author list, affiliations, repository URL and Sustainability (training time / Hardware), using three complementary similarity measures: Python's "difflib.SequenceMatcher" (v3.14.2) (longest common subsequence ratio), Levenshtein ratio (character-level edit distance) and Jaccard similarity (token-set overlap)

### 2.3 Expert curation of the OpenRad repository

Generated model records have been manually verified by 10 expert reviewers (three PhD candidates, 6 MSc candidates and 1 assistant professor, all working on AI research) to ensure correctness and avoid hallucinations. The reviewers assessed whether the extracted information faithfully reflected the source article data and made any necessary corrections, marking them as minor if they involved small inaccuracies like typos or incomplete metrics, or major for critical errors like missing fields or broken links.

### 2.4 Descriptive analysis of the curated repository

The final curated set of JSON files were analysed to reveal underlying trends. This included summarizing the most popular model architectures and validation strategies with bar charts using Seaborn (v0.13.2). Geospatial analysis was also performed by parsing author affiliations. ISO-3 country codes were mapped to organization names using a comprehensive dictionary of over 100 country names, major cities and research institutions. This data was visualized using choropleth maps generated with Plotly (v5.15.0)

Statistical analysis and visualizations were generated using a suite of Python plotting libraries. Matplotlib (v3.7.1) and Seaborn were utilized to create static visualizations, including bar charts for architecture frequency and heatmaps illustrating the cross-distribution of clinical specialties versus imaging modalities. Qualitative insights into model limitations were synthesized using the WordCloud library (v 1.9.6) to identify dominant keywords in the reported limitation sections. Furthermore, a heatmap correlating the reported metrics to model type, such as classification or segmentation, was constructed.

# 3.Results

## 3.1. Platform overview

### 3.1.1. Open-access Model Discovery and Filtering

The main OpenRad dashboard serves as the central hub for model discovery (https://konstvr.github.io/OpenRad/index.html). Users can search for models using keywords or refine their results through a comprehensive set of filters located in the sidebar. These filters include:

- Resource Availability: Finds models for which repositories are available (Github or other), providing open weights or trained models. This data has been collected by reviewing the repositories of all papers, in addition to the publication itself.
- Categorization: Models are tagged by Modality (e.g., CT, MRI, X-Ray), Subspecialty (e.g., Neuroradiology, Chest Radiology, Musculoskeletal) and Use Case (e.g., Classification, Segmentation, Detection). RadLex codes have been used for encoding Modality and Subspecialty, following RSNA roadmap recommendations
- Demo availability: Filter for works that include interactive/downloadable applications

The search results are displayed as a responsive grid of objects (Fig. 2A), each providing a quick summary of the model, including its name and visual badges indicating its verification status, modality, subspecialty and availability of saved weights. This page also includes live and interactiåve overall statistics, regarding the included works, thus summarizing the current landscape of open access AI research (Fig. 2B)

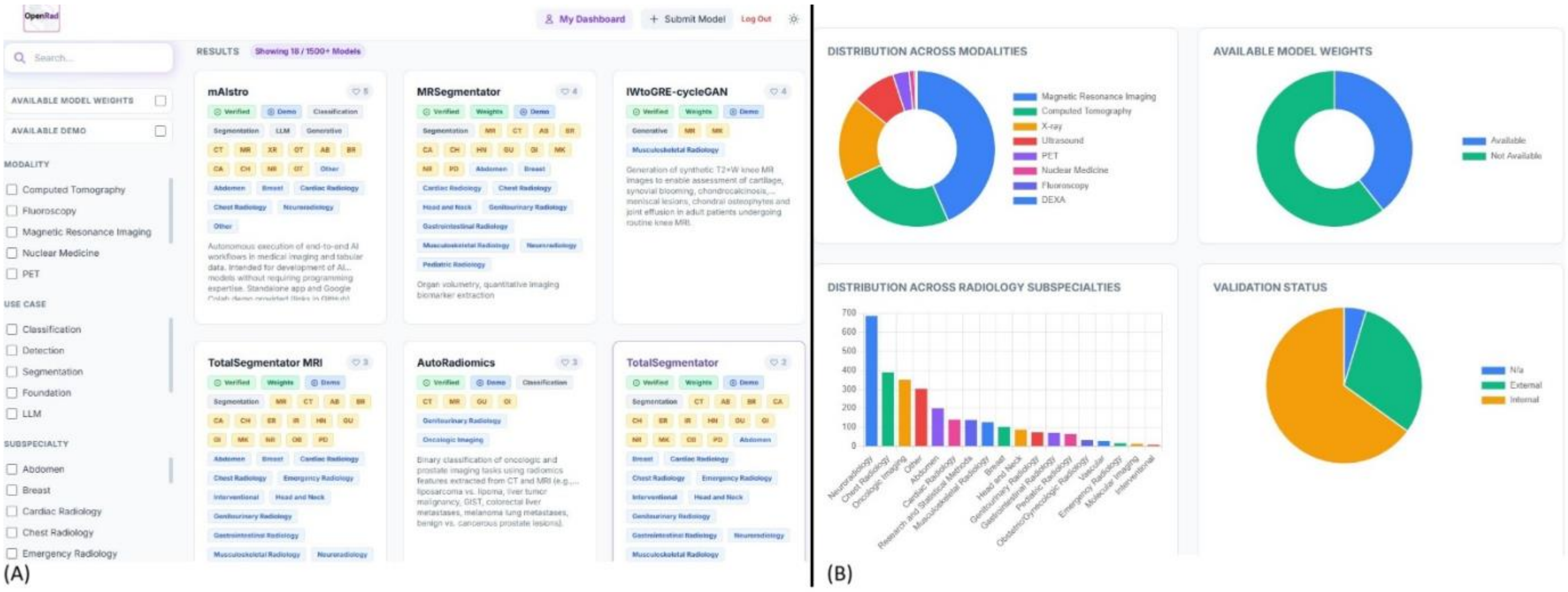

**Fig.2**: *(A): OpenRad Interface. (B) Live, interactive statistics based on the entire model collection*

### 3.1.2. Model record details

Clicking on a model navigates to the detailed view, which presents a structured and comprehensive overview of the model. The interface displays general information, such as the model's name and architecture, alongside technical details regarding the used dataset(s). To assist users in evaluating model suitability, the view includes a section for reported performance metrics and known limitations. Finally, the page serves as a gateway to external resources, providing direct links to code repositories (such as GitHub), the original paper/abstract and hosted interactive demos, when available. For the subset of models (27) with existing ATLAS model cards, that met our inclusion criteria, a link to ATLAS was also included in our model record.

### 3.1.3. Community Contribution and Curation

OpenRad operates on a community-driven model aimed at keeping the registry up-to-date and accurate. Users are encouraged to submit new models to the platform, in addition to the updates done by the authors, ensuring the database continuously expands with the latest research. All community submissions will be verified by experts before being included in the repository, to ensure the validity of submitted data and the consistency of the repository. Beyond submissions, the platform empowers registered users to act as curators. Through a "Verify & Edit" mode, contributors can suggest corrections to metadata or append missing information. These suggestions are tracked and subject to verification to maintain high data quality. Additionally, a flagging system facilitates community moderation, allowing users to report any issues.

## 3.2. Model Card Assessment: Intra-LLM Stability and Human Review

The stability analysis demonstrated an adequate level of consistency across duplicate models for the extracted metadata fields, based on lexical scores (Table 1). Structured content fields related to model identification, including Authors and affiliation list, Repository link, Title and Sustainability showed high stability. Complex, free-text fields such as Model Architecture, Limitations and Regulatory details, exhibited higher variance, reflecting the model's tendency to rephrase descriptive content between generations. Aggregate statistics (mean ± standard deviation) for each metric are reported in Table 2. The qualitative human review confirmed that despite these lexical differences, the technical descriptions exhibited high semantic consistency, indicating that the LLM successfully extracted and synthesized the core meaning of the text. The manual verification showed that for the 1694 included models, 78.5% of edits were classified as minor.

| Field Name | Difflib | Levenshtein Ratio | Jaccard Similarity |
|---|---|---|---|
| **Authors** | 91.3% ± 27.3% | 91.3% ± 27.7% | 90.8% ± 28.5% |
| **Repository link** | 91.4% ± 23.1% | 90.5% ± 25.2% | 87.5% ± 33.0% |
| **Title** | 100.0% ± 0.0% | 100.0% ± 0.0% | 100.0% ± 0.0% |
| **Affiliation list** | 89.5% ± 28.6% | 88.8% ± 29.3% | 89.2% ± 29.2% |
| **Sustainability (training time / Hardware)** | 90.6% ± 22.8% | 88.6% ± 27.0% | 87.0% ± 29.8% |
| **Architecture** | 70.2% ± 35.4% | 66.6% ± 36.2% | 60.8% ± 38.6% |
| **Limitations** | 47.4% ± 44.0% | 50.4% ± 42.2% | 46.0% ± 43.9% |
| **Regulatory** | 79.1% ± 23.4% | 75.4% ± 27.0% | 61.4% ± 40.5% |

**Table 1***: Average Lexical similarity scores and SD for fields with structured content*

### 3.3 Analysis of Repository Data

The analyses shown in Figures 3–7 collectively characterize the current landscape of open access AI models in radiology. Figure 4A displays the distribution of deep-learning architectures: variants of CNN and transformer-based architectures dominate (>33%). Diffusion models are also frequently used (12.1%), followed by U-Net variants and GANs. A smaller number of works utilized YOLO. Figure 4B illustrates the geographic concentration of publications, with China (≈ 400 papers) and the United States (≈ 300 papers) accounting for the majority of the output. Secondary contributions (50-150 papers) came from a limited set of European, Indian and East-Asian nations, while most remaining countries contributed only a handful of studies. Figure 5A presents metric co-occurrence by task type: classification metrics were led by accuracy (229), AUC (166), F1-score (112) and specificity (86). Segmentation accuracy was most commonly measured with DSC (308, which is ≈ 68 % of total DSC mentions), followed by Hausdorff Distance (65) and accuracy (55). Generative tasks primarily used SSIM (88) and PSNR (79). Figure 5B summarizes imaging modality-specialty cooccurrence, highlighting MRI as the most commonly used modality for neuroradiology applications (621 studies), with the next highest volumes observed for X-ray (233) and CT (188) for AI models focusing on chest. Regarding models for Oncologic imaging (OI), MRI (179) and CT (165) were the most utilized modalities. A word-frequency analysis of reported limitations showed that "limited" was the most common term, followed by a cluster of "data, model, dataset, training, performance, image, evaluated". Lower-frequency terms include "small, sample size, external validation, generalization, single center, accuracy," (Supplementary Figure 1)

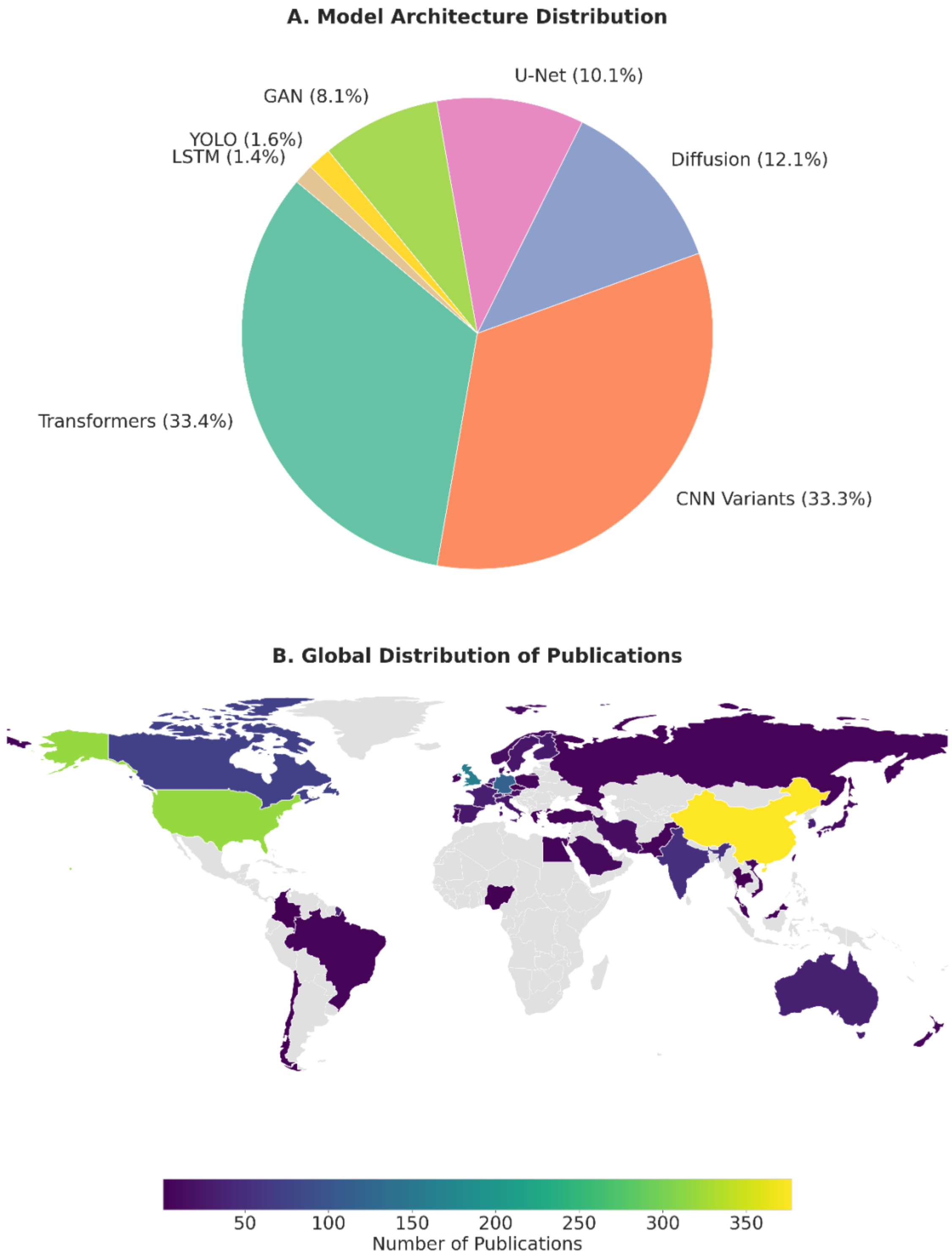

**Fig.4:** *(A): Most frequent model architectures. (B): Global distribution of publications on open access AI models*

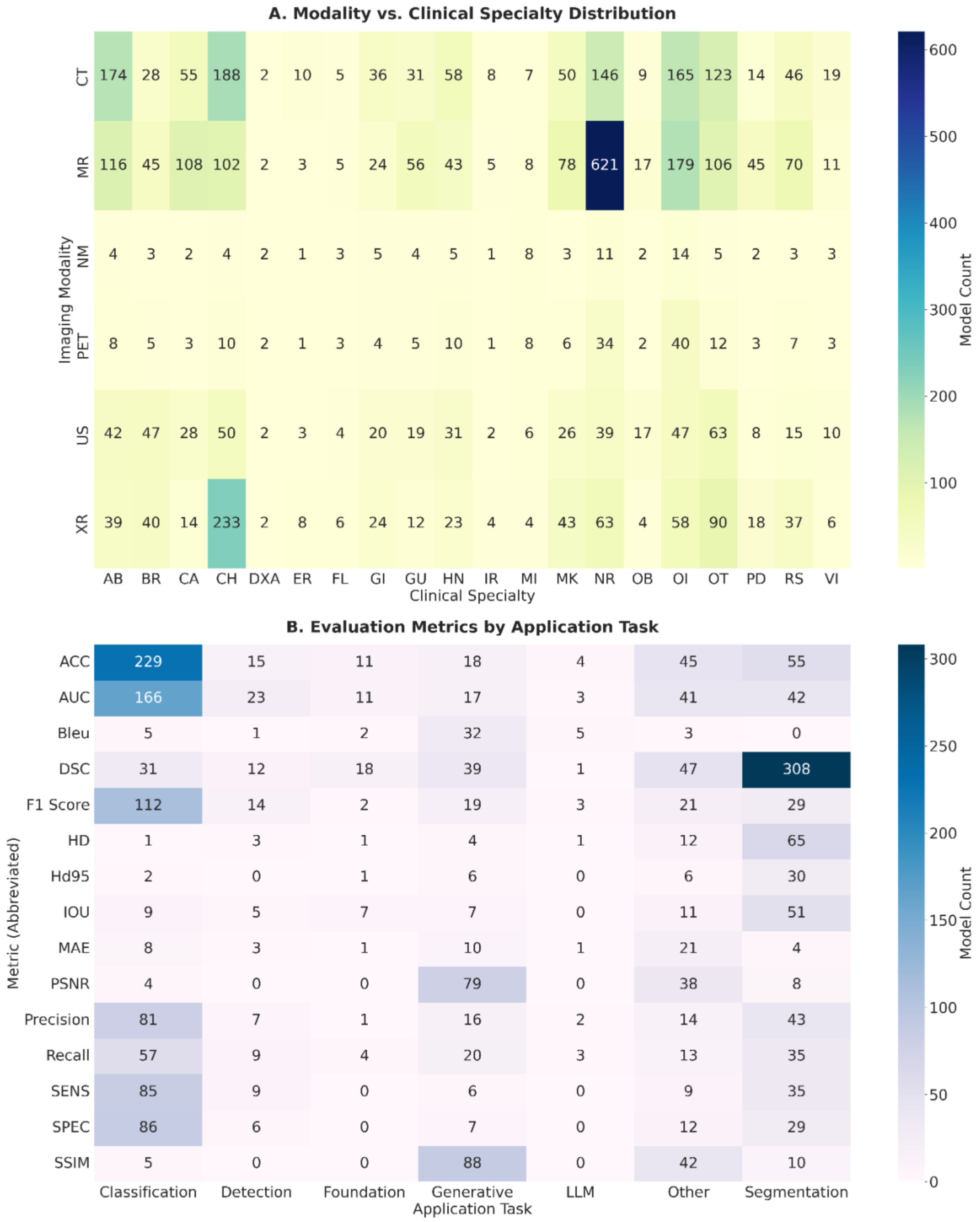

**Fig. 5:** *(A): Distribution of imaging modalities used by AI models across the different radiology subspecialties. Legend: AB: Abdomen, BR: Breast, CA: Cardiac Radiology, CH:*

*Chest Radiology, ER: Emergency Radiology, FL: Fluoroscopy, GI: Gastrointestinal Radiology, GU: Genitourinary Radiology, HN: Head and Neck, IR: Interventional Radiology, MI: Molecular Imaging, MK: Musculoskeletal Radiology, NR: Neuroradiology, OB: Obstetric/Gynecologic Radiology, OI: Oncologic Imaging, OT: Other, PD: Pediatric Radiology , RS: Research and Statistical Methods, VI: Vascular. (B): Distribution of metrics used in the assessment of different model types. Legend: ACC: accuracy, AUC: area under the receiver operating characteristic curve, DSC: dice similarity coefficient, HD: Hausdorff Distance, IOU: intersection over union, MAE: mean absolute error, PCC: Pearson correlation coefficient, PSNR: peak signal-to-noise ratio, SENS: sensitivity, SPEC: specificity, SSIM: structural similarity index measure*

## 4.Discussion

This study introduces the largest publicly accessible, curated repository that aggregates 1694 works on radiology-AI models. By coupling bibliographic metadata with a systematic inspection of the linked code repositories, the platform supplies not only the descriptive information traditionally reported in other databases and research manuscripts, but also an objective appraisal of whether code, pretrained weights or ready-to-use demos are actually available. This dual-layer documentation has the potential to accelerate the transition from literature review to AI implementation, by retrieving code, saved weights or trained models and in many cases bypassing the complex step of re-training from scratch. The platform further supports manual checks that flag broken links or outdated dependencies, ensuring that the catalogue remains functional over time.

The project overcomes two principal obstacles that have limited the use of published AI models (15). First, by summing up all eligible publications in medical literature but also computer science journals and conference proceedings, extracting key metadata and indexing the resulting model records in a faceted web interface, the platform unifies previously scattered code repositories and papers into a single searchable resource. In addition, every card follows a unified schema which captures modality, subspecialty, intended use, data provenance, performance metrics and regulatory details, thereby providing a machine-readable description that facilitates automated analysis and evaluation. The dataset was further enriched with information that is often not present in the original publications, through an automated pipeline which interrogated each corresponding GitHub repository to verify the existence of pretrained model files and when available, live web demonstrations.

The quantitative analyses of the extracted data suggest that publication output remains heavily concentrated in China and the United States, which coincides with rapid technological advances in AI in those geographical areas(16). Architectural preferences have shifted toward CNNs and transformer-based models. Modality-specialty pairings reveal a dominant focus on using MRI data for neuroradiology AI applications (>600 studies) followed by models for chest CT and X-ray, which confirms previous analyses (17). On the contrary, nuclear-medicine and PET applications are scarce. In the word-cloud analysis of reported limitations, the high frequency of the keywords "external validation," "evaluated" and "generalization" alongside "single" and "cohort", highlights a critical gap in robust model validation beyond training populations, with many studies confined to single-center retrospective designs that compromise generalizability across diverse patient populations and imaging protocols.

When compared with existing resources, our work offers a distinct combination of scale, automation and sustainability. The “MedicalModelLibrary” provides a manually curated list of models (11) but is limited to a few entries and focuses mostly on LLMs and Vision models. The RSNA Atlas contains high-quality, expert-curated cards (10), but its coverage is also smaller than the present collection and it does not contain a systematic analysis of code repositories ensuring availably of code, model weights or functional implementations of the AI models (demos). The “Awesome-AI-LLMs-in-Radiology” list aggregates references to LLM applications or report generation (12), but does not deliver structured model records nor any health-check of the associated repositories. By contrast, our platform integrates an automated LLM-driven extraction pipeline, continuous integrity monitoring and a searchable faceted interface, thereby delivering a comprehensive and up-to-date resource for the community.

The study has several limitations. Firstly, model records were generated automatically with an open-source LLM and although stability and fidelity were quantified and all the records were manually reviewed by experts, the process can still produce omissions or minor inaccuracies, especially in cases where the original publication is locked or otherwise inaccessible. Users are therefore encouraged to flag incomplete or erroneous entries and future releases will incorporate the human-curated corrections using the dedicated flagging functionality of the repository. Flagged models are isolated and reviewed by the database curators prior to going live again in an updated version. Secondly, the underlying literature search was restricted to PubMed, arXiv and Scopus using queries defined in the Supplementary Material to ensure reproducibility. While these queries were designed to be comprehensive, they may not have captured every available model in the literature and users are therefore encouraged to enrich the repository through new submissions. Finally, due to computational constraints the initial extraction did not populate every field mandated by the RSNA Roadmap, focusing more on correct indexing and access to manuscript and code.

## 5. Conclusion

OpenRad delivers a large-scale, continuously curated repository of more than 1500 radiology AI models that are indexed, standardized and enriched with verified information on code availability, pretrained weights and live demos. It thus aims to resolve the longstanding problems of fragmented distribution and non-standard reporting, accelerate model discovery and reproducibility, while providing a transparent foundation for future benchmarking and clinical translation of AI in radiology.

## 6.References

**Supplementary Table 1:** *Queries used to create the dataset*

| Database | Query |
|---|---|
| PubMed | ("radiology"[Title/Abstract] OR "imaging"[Title/Abstract]) AND ("github"[Title/Abstract] OR "open access"[Title/Abstract]) AND ("AI"[Title/Abstract] OR "artificial intelligence"[Title/Abstract] OR "ML"[Title/Abstract] OR "machine learning"[Title/Abstract] OR "model"[Title/Abstract]) |
| arXiv | (radiology OR imaging) AND (github OR "open access") AND (AI OR "artificial intelligence" OR ML OR "machine learning" OR model) AND submittedDate:[201512080000 TO 202512072359] |
| Scopus | (TITLE-ABS("radiology" OR "imaging") AND TITLE-ABS("github" OR "open access") AND TITLE-ABS("AI" OR "artificial intelligence" OR "ML" OR "machine learning" OR "model")) AND PUBYEAR > 2014 |

**Supplementary Figure 1:** *Wordcloud of most common words found in limitation sections*

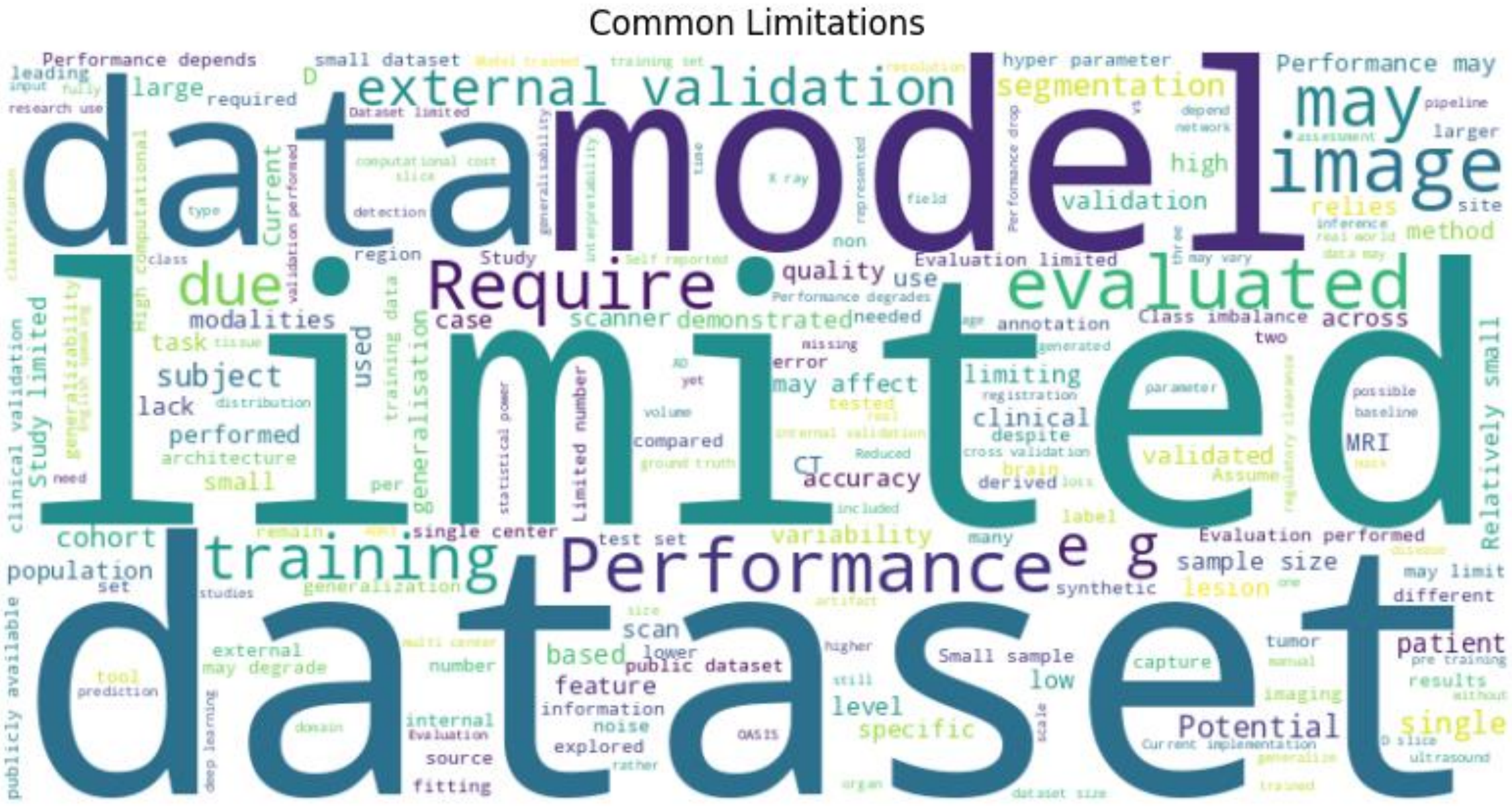